%% file: main.tex
\documentclass[letterpaper]{article} % DO NOT CHANGE THIS
\usepackage{aaai25}  % DO NOT CHANGE THIS
\usepackage{times}  % DO NOT CHANGE THIS
\usepackage{helvet}  % DO NOT CHANGE THIS
\usepackage{courier}  % DO NOT CHANGE THIS
\usepackage[hyphens]{url}  % DO NOT CHANGE THIS
\usepackage{graphicx} % DO NOT CHANGE THIS
\usepackage{natbib}  % DO NOT CHANGE THIS AND DO NOT ADD ANY OPTIONS TO IT
\usepackage{caption} % DO NOT CHANGE THIS AND DO NOT ADD ANY OPTIONS TO IT
\usepackage{algorithm}
\usepackage{algorithmic}

\usepackage{multirow}

\usepackage{booktabs} 

\usepackage{newfloat}
\usepackage{listings}

\DeclareCaptionStyle{ruled}{labelfont=normalfont,labelsep=colon,strut=off} % DO NOT CHANGE THIS
\floatstyle{ruled}
\newfloat{listing}{tb}{lst}{}
\floatname{listing}{Listing}
\title{DDD: Discriminative Difficulty Distance for plant disease diagnosis}
\author{
    Yuji Arima \textsuperscript{\rm 1}, 
    Satoshi Kagiwada \textsuperscript{\rm 2}, 
    Hitoshi Iyatomi\textsuperscript{\rm 1}
}
\affiliations{
    \textsuperscript{\rm 1} Graduate School of Science and Engineering Major in Applied Informatics, Hosei University, Tokyo, Japan \\
    \textsuperscript{\rm 2 } Clinical Plant Science, Faculty of Bioscience and Applied Chemistry, Hosei University, Tokyo, Japan \\
    yuji.arima.3m@stu.hosei.ac.jp, 
    kagiwada@hosei.ac.jp, 
    iyatomi@hosei.ac.jp
}

\usepackage{bibentry}

\usepackage{diagbox}
\usepackage {tabularray}
\usepackage{amsmath} 
\usepackage{threeparttable}
\usepackage{amsmath}
\usepackage{amssymb}
\usepackage{amsfonts}
\usepackage{bm}

\begin{document}
\maketitle

\begin{abstract}
    \input{00_abstract}
\end{abstract}

\section{Introduction}
    \input{01_introduction}

\section{Related Works}
    \input{02_related_works}

\input{tables/01_Datasets/datasets}

\section{Discriminative Difficulty Distance (DDD)
}
    \input{03_method}

\input{images/01_Result/result}
\input{tables/02_Result/result}

\section{Experiments}
    \input{04_experiments}

\input{images/02_Parameter/parameter}

\section{Result}
    \input{05_result}

\section{Discussion}
    \input{06_discussion}

\input{tables/03_Result_of_detail/result_of_detail}

\section{Limitation of study}
    \input{07_limitation}

\section{Conclusion}
    \input{08_conclusion}

\section{Acknowledgments}
    \input{09_Acknowledgments}

\bibliography{reference}

\end{document}

%% file: 00_abstract.tex
Recent studies on plant disease diagnosis using machine learning (ML) have highlighted concerns about the overestimated diagnostic performance due to inappropriate data partitioning, where training and test datasets are derived from the same source (domain). Plant disease diagnosis presents a challenging classification task, characterized by its fine-grained nature, vague symptoms, and the extensive variability of image features within each domain. In this study, we propose the concept of Discriminative Difficulty Distance (DDD), a novel metric designed to quantify the domain gap between training and test datasets while assessing the classification difficulty of test data. DDD provides a valuable tool for identifying insufficient diversity in training data, thus supporting the development of more diverse and robust datasets. We investigated multiple image encoders trained on different datasets and examined whether the distances between datasets, measured using low-dimensional representations generated by the encoders, are suitable as a DDD metric. The study utilized 244,063 plant disease images spanning four crops and 34 disease classes collected from 27 domains. As a result, we demonstrated that even if the test images are from different crops or diseases than those used to train the encoder, incorporating them allows the construction of a distance measure for a dataset that strongly correlates with the difficulty of diagnosis indicated by the disease classifier developed independently. Compared to the base encoder, pre-trained only on ImageNet21K, the correlation higher by 0.106 to 0.485, reaching a maximum of 0.909.

%% file: 01_introduction.tex
Numerous machine learning (ML)-based diagnostic models for plant diseases and pests have been proposed, achieving impressive numerical results in various studies ~\cite{plant_mohanty,plant_atila2021plant,plant_elfatimi2022beans,plant_Fujita2018APP,plant_hughes2015open,plant_kawasaki2015basic,plant_narayanan2022banana,plant_ramcharan2017deep,plant_toda2019convolutional,plant_wang2017automatic}. However, many of these models were evaluated on improper data partitioning practices, wherein the training and validation datasets originate from the same photographing environment. This improper data partitioning has led to overestimation of diagnostic performance, as recent studies have highlighted that the actual diagnostic accuracy of even state-of-art models is significantly lower than that reported ~\cite{ppig_plant_9398772, HODRF_plant_iwano2024hierarchicalobjectdetectionrecognition,leafgan_plant_plant_cap2020leafgan,21k_plant_shibuya2021validation}. Plant disease diagnosis represents a highly challenging fine-grained classification task, as the diagnostic cues —disease symptoms— are often ambiguous and subtle. The significant image diversity caused by variations in composition, background, plant variety, disease progression, and other domain-specific factors, further complicates the task. Moreover, disease lesions frequently occupy a small portion of the overall image, making it challenging for ML models to generalize effectively. When the variety of the training data is limited, as observed in previously reported studies, ML models tend to overfit to a narrow set of training patterns, resulting in suboptimal generalization performances. This inherent limitation arises from the inability of the training data to adequately capture the diversity of classification targets. In particular, constructing a diagnostic model capable of achieving high classification performance on data with unseen characteristics is especially difficult when there is significant variation between the domains of the training and evaluation datasets. For instance, in a plant disease diagnosis study by Shibuya et al., utilizing a large dataset of more than 221,000 images spanning four crops ~\cite{21k_plant_shibuya2021validation}, even the utilization of advanced classification algorithms failed to achieve satisfactory accuracy for the diseases exhibiting significant domain gaps. The result of that study underscores the critical need for robust methodologies to address domain variability and enhance the diagnostic capabilities of ML models for plant diseases.

In most ML tasks, the critical challenge consists of obtaining a low-dimensional representation that captures domain-independent and task-relevant features. Two well-established methods for achieving this goal are metric learning and contrastive learning. Both methods aim to train models that bring the low-dimensional representations of data from the same class closer together while pushing apart representations of data from different classes. Metric learning is a supervised learning approach that utilizes class labels to guide the representation learning process. In contrast, researchers often implement contrastive learning as a self-supervised learning method. It relies on constructing positive and negative pairs, typically through data augmentations or clustering-based pseudo-labeling, to learn representations without explicit class annotations. Models employing contrastive learning have demonstrated notable success in tasks characterized by significant domain variations, showcasing their effectiveness in depicting generalizable features across diverse datasets.

However, when dealing with problems that have a narrow range of difficulty and a significant domain gap between training and test data, limited training data becomes insufficient for effective harmonization, rendering it ineffective in bridging the gap. In such cases, one may relax the problem into a typical semi-supervised learning framework (i.e., transductive learning), where the model observes only the test data, or adopt a setting where both the data and labels are visible, albeit only for a portion of the test data. The former approach has shown success in a variety of general ML tasks ~\cite{Self-training}. However, it is less effective in domains such as plant disease diagnosis, where there are substantial discrepancies in the training and test data trends because the accuracy of pseudo-teacher labels estimated for the test data is poor. On the other hand, the latter approach provides a more realistic solution for fine-grained problems exhibiting significant domain gaps. Key research questions in this context include determining the minimal number of data points required to achieve a specific performance level for each task and optimizing the use of the limited available data. Ultimately, the success of these approaches heavily depends on the magnitude of the domain gap. Therefore, from a ML perspective, quantifying the size of the domain gap is crucial. Although some previous studies have attempted to measure the distance between datasets, they have only investigated the transferability of knowledge by measuring the distance between datasets, and more research is needed on indicators to guide effective measures for addressing problems with fine-grained domain gaps.

In light of this context, this paper proposes the concept of Discriminative Difficulty Distance (DDD) as a novel metric for quantifying the domain-gap differences between image datasets in fine-grained tasks. The paper also reports on initial investigations into practical methods for the application of DDD, with a particular focus on plant disease diagnosis. The proposed DDD is a pseudo-measure, designed to quantitatively capture the divergence between training and test datasets to indicate the difficulty in diagnosing test data using a ML model trained on the training dataset. A large DDD between the training and test datasets suggests that the training data lacks sufficient diversity, highlighting an opportunity for intervention, such as adding or generating more training data. Furthermore, when constructing a training dataset using data from multiple domains, DDD can be valuable for understanding the diversity within each domain, thereby facilitating the creation of a more varied and robust training dataset. Additionally, the DDD for each classification label may offer insights into the difficulty of the classification task, providing a helpful trigger for taking appropriate action. When implemented effectively, the DDD has the potential to substantially contribute to developing more robust ML models.

In this paper, we evaluate the validity of $S$ (distance $L$) as an indicator of DDD by calculating the similarity between two datasets obtained using multiple ML models in the plant disease diagnosis task. Specifically, we calculate the confusion matrix $P$ of the classification results of different datasets using the discriminator trained on one dataset, and measure the correlation between $S$ and $P$ as an evaluation metric.

The main contributions of this paper can be summarized as follows.
\begin{itemize}
\item[1)] The proposal of the DDD as a novel metric for quantifying the domain gap between datasets, specifically in terms of the difficulty associated with classification tasks, along with a detailed method for its calculation.
\item[2)] In the plant disease diagnosis task, it was discovered that image encoders trained on images of crops other than the target crop can also be a powerful method for obtaining DDD.
\end{itemize}

%% file: 02_related_works.tex
\subsection{Metric learning}
Metric learning has emerged as a prominent research area in recent years, particularly in computer vision. Computer vision has gained significant attention due to the considerable variation in visual features observed among data samples from the same class. Wang et al. introduced the Triplet loss ~\cite{metric_learning_triplet}, a loss function designed to learn similarity distances from images directly. The main goal of this approach is to construct a separable embedding space that effectively captures subtle visual similarities between images within the same category. Building on this foundation, Sohn et al. expanded the Triplet loss by introducing the multi-class N-pair loss 
 ~\cite{metric_learning_n-class}, offering a more flexible framework for learning representations. Additionally, Chen et al. developed baseline++ ~\cite{metric_learning_baseline++}, a method that reduces intra-class variation by leveraging the cosine distance between input features and the weight vectors of each class. These approaches share a common focus on designing loss functions to optimize feature space distances, thereby enhancing discrimination and improving model performance.

\subsection{Contrastive Learning}
Researchers have widely adopted contrastive learning to automate the expensive manual annotation of large amounts of data and to investigate the relationships between different forms of data for multi-modal applications. In addition, contrastive learning has achieved state-of-the-art results through its discriminative learning framework. Chen et al. introduced SimCLR ~\cite{contrastive_learning_SimCLR}, a self-supervised learning approach that integrates certain supervised elements. Chechik et al. trained a large-scale image similarity model for retrieval using triplet loss ~\cite{metric_learning_triplet}, based on the concept of invariant mapping and its application to metric learning ~\cite{contrastive_learning_10.1007/978-3-642-02172-5_2}. Similarly, Radford et al. developed CLIP ~\cite{contrastive_learning_CLIP}, which utilizes contrastive learning to model the relationship between images and text. These contrastive learning methods focus on the distances within the latent space, similar to distance learning, and these distances are optimized and used as part of the learning objective to improve the representation.

\subsection{Distance between datasets}
Measuring the distance between datasets is an essential indicator of the transferability of knowledge. Calderon-Ramirez et al. introduced DeDiM ~\cite{dataset_distance_DeDiM}, a metric designed to assess dataset similarity to evaluate semi-supervised learning in the context of distributional discrepancies between labeled and unlabeled datasets. Similarly, Alvarez-Melis et al. developed OTDD ~\cite{dataset_distance_OTDD}, a method that utilizes optimal transport to compute the distance between datasets. While both approaches effectively quantify the distance between datasets,  they fall short in providing targeted improvements for tasks that require handling fine-grained features or addressing significant domain gaps.

%% file: tables/01_Datasets/datasets.tex
\begin{table*}[t]
\caption{DETAILS OF THE DATASET. }  
\label{tb:dataset}
\input{tables/01_Datasets/datasets/cucumber} \hfill
\input{tables/01_Datasets/datasets/tomato} \hfill
\input{tables/01_Datasets/datasets/eggplant} \hfill
\input{tables/01_Datasets/datasets/strawberry}
\end{table*}

%% file: tables/01_Datasets/datasets/cucumber.tex
\begin{minipage}[t]{.47\textwidth}
\begin{center}
\begin{tabular}{lrr}
\multicolumn{3}{c}{cucumber} \\ 
\toprule
{ID\_Name} & \multicolumn{1}{l}{{train}} & \multicolumn{1}{l}{{test}}  \\ 
\midrule
00\_HEALTHY                  & 16,023    & 5,576 \\
01\_Powdery\_Mildew          & 7,764     & 1,898 \\
02\_Gray\_Mold               & 643       & 167   \\
03\_Anthracnose              & 3,038     & 77    \\
08\_Downy\_Mildew            & 6,953     & 2,579 \\
09\_Corynespora\_Leaf\_Spot  & 7,565     & 1,813 \\
17\_Gummy\_Stem\_Blight      & 1,483     & 374   \\
20\_Bacterial\_Spot          & 4,362     & 2,648 \\
22\_CCYV                     & 5,969     & 179   \\
23\_Mosaic\_diseases         & 26,861    & 1,626 \\
24\_MYSV                     & 17,239    & 1,004 \\
\midrule
Total                        & 97,900    & 17,941  \\
\bottomrule
\end{tabular}
\end{center}
\end{minipage}

%% file: tables/01_Datasets/datasets/tomato.tex
\begin{minipage}[t]{.47\textwidth}
\begin{center}
\begin{tabular}{lrr}
\multicolumn{3}{c}{tomato} \\ 
\toprule
{ID\_Name} & \multicolumn{1}{l}{{train}} & \multicolumn{1}{l}{{test}}  \\ 
\midrule
00\_HEALTHY                    & 8,120     & 2,994 \\
01\_Powdery\_Mildew            & 4,490     & 4,250 \\
02\_Gray\_Mold                 & 9,327     & 571   \\
05\_Cercospora\_Leaf\_Mold     & 4,078     & 1,809 \\
06\_Leaf\_Mold                 & 2,761     & 151   \\
07\_Late\_Blight               & 2,049     & 808   \\
10\_Corynespora\_Target\_Spot  & 1,732     & 1,350 \\
19\_Bacterial\_Wilt            & 2,259     & 412   \\
21\_Bacterial\_Canker          & 4,369     & 128   \\
27\_ToMV                       & 3,453     & 49    \\
28\_ToCV                       & 4,320     & 871   \\
29\_Yellow\_Leaf\_Curl         & 4,513     & 1,746 \\
\midrule
Total                       & 51,471    & 15,139  \\
\bottomrule
\end{tabular}
\end{center}
\end{minipage}

%% file: tables/01_Datasets/datasets/eggplant.tex
\begin{minipage}[t]{.47\textwidth}
\begin{center}
\begin{tabular}{lrr}
\multicolumn{3}{c}{eggplant} \\ 
\toprule
{ID\_Name} & \multicolumn{1}{l}{{train}} & \multicolumn{1}{l}{{test}}  \\ 
\midrule
00\_HEALTHY               & 12,431    & 1,122 \\
01\_Powdery\_Mildew       & 7,936     & 938   \\
02\_Gray\_Mold            & 1,024     & 166   \\
06\_Leaf\_Mold            & 3,188     & 732   \\
11\_Leaf\_Spot            & 5,510     & 118   \\
18\_Verticillium\_Wilt    & 3,176     & 354   \\
19\_Bacterial\_Wilt       & 3,415     & 462   \\
\midrule
Total                     & 36,680    & 3,892  \\
\bottomrule
\end{tabular}
\end{center}
\end{minipage}

%% file: tables/01_Datasets/datasets/strawberry.tex
\begin{minipage}[t]{.47\textwidth}
\begin{center}
\begin{tabular}{lrr}
\multicolumn{3}{c}{strawberry} \\ 
\toprule
{ID\_Name} & \multicolumn{1}{l}{{train}} & \multicolumn{1}{l}{{test}}  \\ 
\midrule
00\_HEALTHY             & 10,472    & 578   \\
01\_Powdery\_Mildew     & 1,952     & 893   \\
03\_Anthracnose         & 3,701     & 609   \\
15\_Fusarium\_Wilt      & 2,608     & 227   \\
\midrule
Total                   & 18,733    & 2,307  \\
\bottomrule
\end{tabular}
\end{center}
\end{minipage}

%% file: 03_method.tex
\subsection{Significance}
In this paper, we propose the concept of Discriminative Difficulty Distance (DDD) as a new metric for objectively quantifying the differences between image data or image datasets using ML models. DDD is a pseudo-distance metric between datasets calculated based on low-dimensional representations (embeddings) of data. It aims to quantify the classification difficulty of one dataset relative to another, as perceived by a ML model trained on the first dataset. As a specific application, we explore effective implementation strategies for plant disease diagnosis, a fine-grained task characterized by a significant domain gap. Calculating the DDD between the training and evaluation datasets serves as a crucial measure for assessing the diversity of the training dataset and aids in identifying challenging-to-classify classes. Moreover, it facilitates the construction of a more desirable, broader, and robust training dataset necessary for the task. 

\subsection{Implementation Policy}
In this paper, we consider a more appropriate method for calculating DDD. 
Specifically, we propose that the distance between two datasets, calculated using low-dimensional representations generated by a suitably designed image encoder ($M_E$), serves as a viable candidate for DDD. We further evaluate its effectiveness in this same context.
When we assume that the ML model $M_C$ has been adequately trained for the target task, its ability to correctly identify disease $a$ with high probability suggests that diagnosing disease $a$ is relatively straightforward for $M_C$. This implies that the characteristics of disease $a$ in the training data are similar to those in the test data.
Conversely, if $M_C$ frequently miss-classifies disease $a$ as disease $b$, it suggests that the characteristics of disease $a$ in the training data are similar to those of disease $b$ in the test data.
Based on this assumption, we can verify whether the diagnostic similarity $S_{ij}$ between disease $i$ in one dataset and disease $j$ in the other, as calculated using $M_E$, reflects the confusion matrix $P_{ij}$ which can be calculated from the estimated probability of data classification by the ML model. If these values are highly consistent, we can determine that the distance calculated using the low-dimensional representation constructed by $M_E$ is an effective DDD.

\subsection{Implementation and evaluation of a reasonable distance as DDD in the plant disease diagnosis task}
Let $X_t (t={\mathrm{train}, \mathrm{test}})$ be the dataset to be used to calculate the similarity between datasets, $M_E$ be the image encoder that converts images into low-dimensional representations, and $M_C$ be the plant disease identifier constructed by training on a sufficiently large $X_{\mathrm{train}}$ (a total of 244,063 images in Table 1 below). The objective is to devise a construction strategy for the image encoder $M_E$ that yields a dataset distance closely aligned with the probability distribution of the predictions made by $M_C$ for each disease in $X_{\mathrm{test}}$.
The following are the steps we propose for implementation.

\subsubsection{(step 1) Acquisition of a low-dimensional representation for each dataset}
Each data in the datasets $X_{\mathrm{train}}$ and $X_{\mathrm{test}}$ is converted into a low-dimensional representation, $\bm{z_i}$ and $\bm{z_j}$, using an image encoder $M_E$ (e.g., a CNN model pre-trained on ImageNet21k~\cite{ridnik2021imagenet21k}). Here, $i$, $j$ represent the indices of the classification classes (i.e., disease types) of the training and test data, respectively ($i, j = 1, 2,\cdots, C$). The number of data in classes $i$ and $j$ are represented as $N_i$ and $N_j$, respectively.

\subsubsection{(step 2) Calculation of the average class vectors of test data}
Compute the mean vector 
$$\bm{\bar{z}_{j}} = \frac{1}{N_j}\sum_{k=1}^{N_j}\bm{{z}_{jk}} $$ 
of the low-dimensional representation of each data in class $j$ of the test dataset.

\subsubsection{(step 3) Calculation of the diagnostic distance $L_{ij}$ and the diagnostic similarity $S_{ij}$}
The  diagnostic distance $L_{ij}$ between each vector for class $i$ in the training dataset and class $j$ in the test dataset is computed as follows: 
$$L_{ij} = \frac{1}{N_i}\sum_{l=1}^{N_i}||\bm{z_{il}}-\bm{\bar{{z}}_{j}}|| .$$ 
The diagnostic similarity is defined as
$$S_{ij}=\frac{\exp{(-\alpha L_{ij})}}{\sum^{C}_{m=1}{\exp{(-\alpha L_{mj})}}},$$ 
where $\alpha$ is an adjustable hyperparameter. $S_{ij}$ is calculated based on the diagnostic distance $L_{ij}$ and is the similarity between class $i$ of the training dataset and class $j$ of the test data in the range [0,1].

\subsubsection{(step 4) Validation of diagnostic similarity $S$}
Based on our hypothesis described in the "Implementation Policy" section, we will verify whether the diagnostic similarity $S$ obtained using $M_E$ reflects the difficulty of data classification by the ML model. First, we use the classifier $M_C$ built using the training data $X_{\mathrm{train}}$ to obtain the confusion matrix $P$ when the test data is diagnosed. Each element $P_{ab}$ of the confusion matrix $P$ represents the proportion of instances where disease $a$ is miss-classified as $b$. The values are normalized such that the total proportion of all classifications for disease a sums to 1. At this time, $P_{a=i, b=j}$ and the diagnostic similarity $S_{ij}$ are compared across all combinations of ${i,j}$, and the correlation $R$ is calculated as 
$$ R = r(\bm{\Tilde{P}},\bm{S}),$$ 
where $\bm{\Tilde{P}} =\{P_{a=i, b=j}\}$ and $\bm{S}=\{{S_{ij}}\} \in \mathbb{R}^{C^2}$. The function $r()$ calculates the similarity between two vectors, and cosine similarity was utilized in this experiment.
Importantly,this correlation accounts not only for the model's accuracy but also for the patterns in its error-making tendencies. 

\subsection{Notes}
Suppose $M_C$ and $M_E$ use the same architecture and share the training data. In that case, $R$ is necessarily higher, as the low-dimensional representations obtained for a given image will be very similar for both $M_E$ and $M_C$. Therefore, ensuring that the training data for $M_E$ differs from that of $M_C$ is crucial. For comparison and discussion, we also evaluated under these conditions in this experiment. These results are marked with \textdagger \, in the Results column, as described below.

%% file: images/01_Result/result.tex
 \begin{figure*}[ht]
 \begin{center}
  \includegraphics[width=0.71\linewidth]{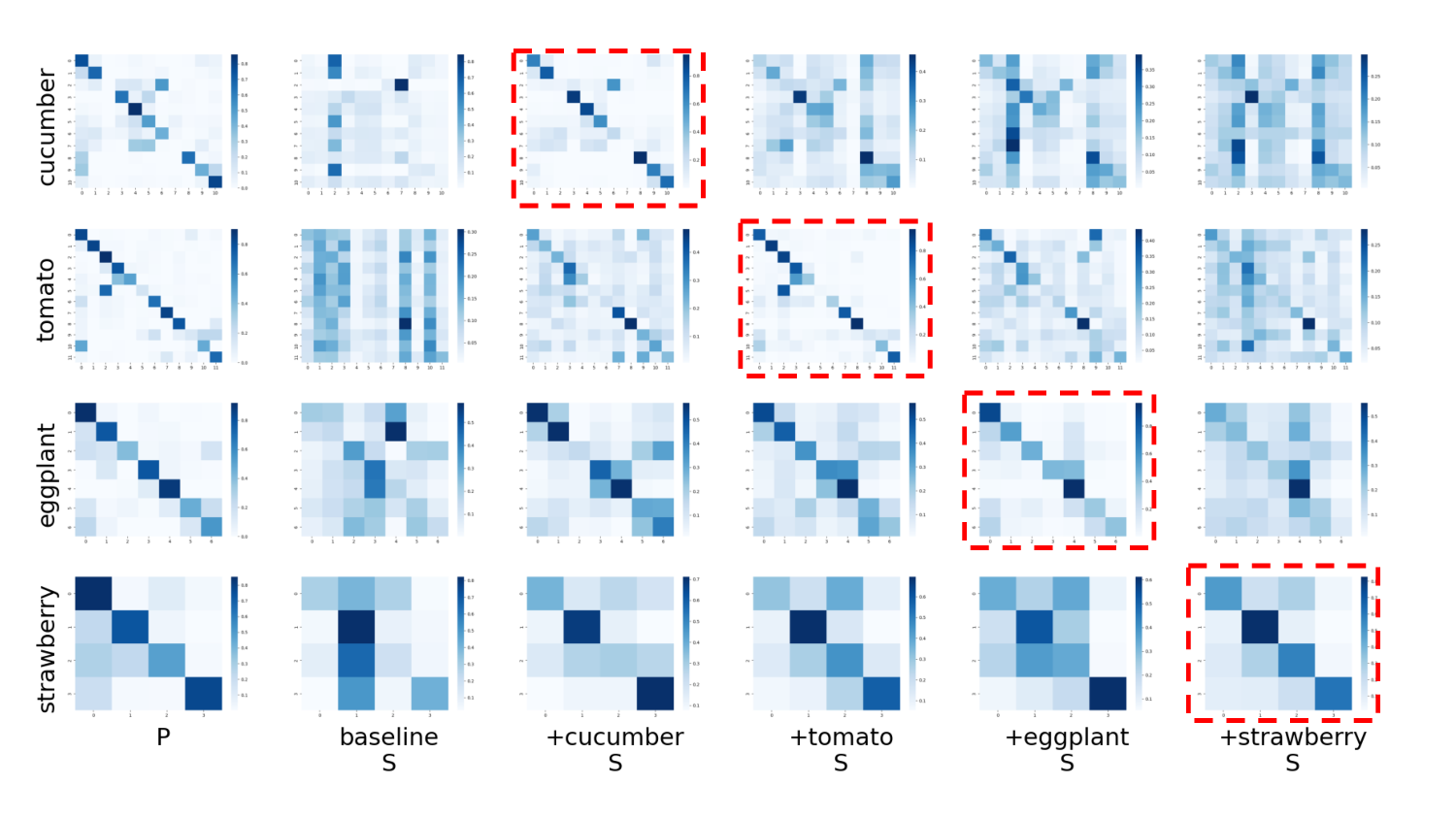}
  \caption{Comparison of the confusion matrix ($P$: left-most column) for each crop diagnosis by $M_C$ and the diagnostic similarity ($S_{ij}$: remaining columns) between both data sets generated by each $M_E$. Dark-color indicate high values. The results in the dashed boxed area are for reference only, as the part of training data for $M_C$ and $M_E$ are shared.}
  \label{fig:result}
 \end{center}
\end{figure*}

%% file: tables/02_Result/result.tex
\begin{table*}[htbp]
\centering
\caption{Summary of correlations ($R$) between the probability of disease diagnosis by $M_C$ ($\tilde{P}$) and the Diagnostic similarity ($S_{ij}$: remaining columns) between both data sets estimated by each $M_E$ (at $\alpha=1.0$).}
\label{tb:result}

\begin{tabular*}{0.62\linewidth}{l|rrrr}
\hline
\diagbox[height=2\line]{$M_E$}{Target Crop} 
& cucumber  & tomato & eggplant & strawberry \\ 
\hline
baseline        & 0.232           & 0.468           & 0.544           & 0.714  \\
\quad+cucumber   & 0.944\TblrNote{\textdagger}      & 0.743           & \textbf{0.909}  & \textbf{0.896} \\
\quad+tomato     & \textbf{0.717} & 0.966\TblrNote{\textdagger}       & 0.876           & 0.864   \\
\quad+eggplant   & 0.564          & \textbf{0.745}  & 0.932\TblrNote{\textdagger}       & 0.868   \\
\quad+strawberry & 0.553          & 0.574           & 0.704           & 0.925\TblrNote{\textdagger}  \\
\hline
% \end{talltblr}
\multicolumn{5}{l}{${}^{\dagger}$ The results are for reference only, as the training data for $M_E$ includes} \\
\multicolumn{5}{l}{\quad  some of the training data for $M_C$.} \\
\multicolumn{5}{l}{Bolded values indicate the best values excluding reference results.}

\end{tabular*}

\end{table*}

%% file: 04_experiments.tex
\subsection{Dataset}
Table \ref{tb:dataset} shows an overview of the training and test data, which consists of 244,063 leaf images from 34 classes of the four crops used in this experiment.
Between 2016 and 2020, experts individually cultivated the plants, inoculated them with pathogens, photographed them, and labeled the images under strict disease control at 27 agricultural experiment stations across 24 prefectures in Japan. The images generally focus on a single leaf near the center, although many also feature multiple leaves at various distances from the subject.
In our experimental setup, we separated the training and test images by ensuring they came from different locations, and we strictly evaluated the test data as entirely unknown.

\subsection{Implementation}
In this study, we adopted EfficientNetV2-m (EfficientNetV2) ~\cite{tan2021efficientnetv2}, which was pre-trained on ImageNet21k ~\cite{ridnik2021imagenet21k}, for $M_C$ and $M_E$ with primitive data augmentation techniques including random rotation, flipping, cropping, and rectangular masking for data augmentation. As an optimizer, SGD with a learning rate of $1.0\times10^{-4}$ and a momentum of $0.9$ was used. As will be introduced later, $M_C$ trained all the Training data for all four crops, whereas each of the four $M_E$s trained the training data for one crop.

\subsection{Comparison Methods}
In this study, we discussed and evaluated the optimal configuration of the encoder $M_E$, which acquires low-dimensional representations that calculate the distance between images to achieve DDD. To do this, we compared five models that used the same ML model as mentioned above but were trained on different images. For the evaluation of $M_E$, we used $R$ and $R_d$ defined in “Implementation and evaluation of a reasonable distance as DDD in the plant disease diagnosis task” according to the policy described in "Implementation Policy".

\begin{itemize}
\item[] \textbf{baseline} is pre-trained by large open dataset ImageNet21k.
\item[] \textbf{+cucumber} has been additionally trained using the baseline cucumber training image dataset.
\item[] \textbf{+tomato} has been additionally trained using the baseline tomato training image dataset.
\item[] \textbf{+eggplant} has been additionally trained using the baseline eggplant training image dataset.
\item[] \textbf{+strawberry} has been additionally trained using the baseline strawberry training image dataset.
\end{itemize}

%% file: images/02_Parameter/parameter.tex
 \begin{figure*}[ht]
 \begin{center}
  \includegraphics[width=0.70\linewidth]{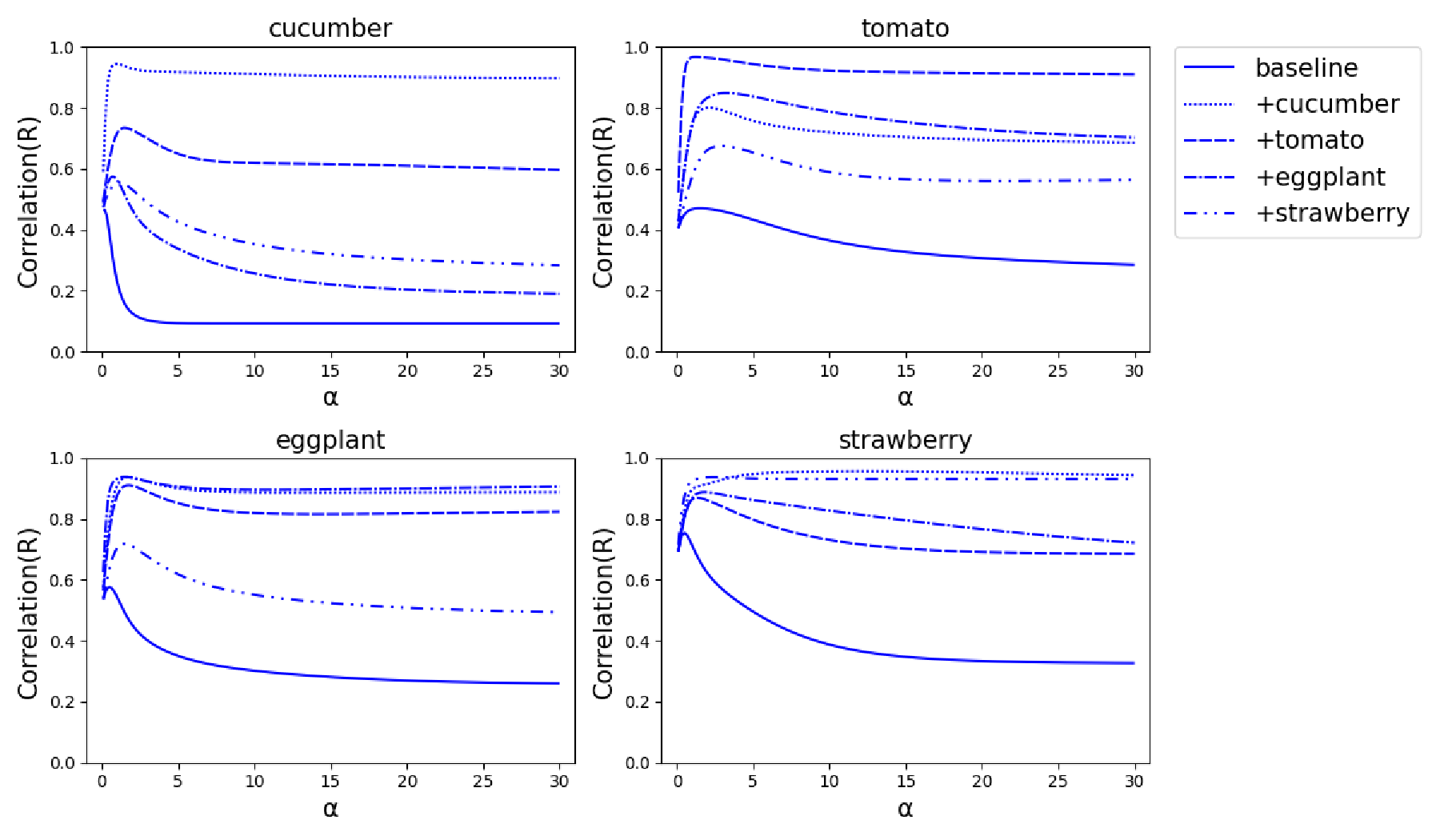}
  \caption{The dependence of correlation $R$ on hyperparameter $\alpha$.}
  \label{fig:parameter}
 \end{center}
\end{figure*}

%% file: 05_result.tex
Figure \ref{fig:result} shows a comparison of the confusion matrix ($P$: leftmost column) for each crop diagnosis by $M_C$ and the diagnostic similarity ($S_{ij}$: remaining columns) between both datasets generated by each $M_E$.
Table \ref{tb:result} is a summary table of the correlation ($R$) between the probability of disease diagnosis by $M_C$ ($\tilde{P}$) and the diagnostic similarity ($S_ij$: remaining columns) between both datasets estimated by each $M_E$. Table \ref{tb:result_of_detail} shows a breakdown of the results from Table \ref{tb:result} by disease. Figure \ref{fig:parameter} shows the dependence of correlation $R$ on hyperparameter $\alpha$. The results marked with \textdagger \, in Tables \ref{tb:result} and \ref{tb:result_of_detail} and the part of Figure \ref{fig:result} enclosed by the dotted line are reference results because the training data for $M_E$ and $M_C$ were shared.

%% file: 06_discussion.tex
\subsection{Validity Evaluation}
Table \ref{tb:result}, \ref{tb:result_of_detail} shows that the $M_E$ fine-tuned for each plant has achieved higher $R$ than the baseline for unknown crops. This result means that in plant disease diagnosis, even for unknown crops, it is possible to obtain low-dimensional representations that show more accurate diagnostic potential than the baseline from the $M_E$ fine-tuned for each plant. These results suggest there are clues to plant image diagnosis other than the universal image features learned using large-scale data. Furthermore, even when a researcher fine-tunes the model on other crops,this knowledge can still be partially leveraged,  serving as a valuable tool that provides an effective indicator for seeking diagnostic potential.

Let us compare the $S$ values for eggplant, obtained by +cucumber, which showed a high $R$ in Figure \ref{fig:result}, to the $P$ values for eggplant alone. The areas with a significant tendency for diagnostic errors, beyond just the diagonal components, align closely. This result suggests that DDD can also help address bottlenecks in diagnostic ability.

\subsection{Parameter Analysis}
Figure \ref{fig:parameter} shows that almost all $M_E$ models reach their maximum $R$ when $\alpha$ is between 0.1 and 5.0. 
This observation suggests that the most suitable similarity measure $S$ for DDD can be achieved by appropriately adjusting $\alpha$. It is worth noting that the $R$ for the +cucumber dataset is exceptionally high in the case of strawberry. Strawberry is a relatively simple task with a smaller domain-gap than other crops. So, the knowledge used to identify the disease class in the cucumber training dataset encompasses much of the knowledge used to identify the disease class in strawberry. As a result, the knowledge from the cucumber training dataset is highly transferable to the strawberry dataset, leading to a higher $R$ for the +cucumber dataset when applied to strawberry.

%% file: tables/03_Result_of_detail/result_of_detail.tex
\begin{table*}[htbp]
\centering
\caption{Details of the correlations between the probability of disease diagnosis by $M_C$ ($\tilde{P}$) and the diagnostic similarity ($S_{ij}$: remaining columns) between both datasets estimated by each $M_E$ (at $\alpha=1.0$).}
\label{tb:result_of_detail}

% header
\begin{tabular}{llrrrrr}

\toprule
% Target Crop & ID\_Name & baseline & +cucumber  & +tomato & +eggplant & +strawberry \\
\multirow{2}{*}{Target Crop \qquad} & \multirow{2}{*}{ID\_Name \qquad\qquad\qquad\qquad\qquad} & \multicolumn{5}{c}{$M_E$} \\
  &   & baseline & +cucumber  & +tomato & +eggplant & +strawberry \\
\midrule

% cucumber
\multirow{12}{*}{cucumber} &
00\_HEALTHY                   & 0.069     & 0.987${}^{\dagger}$   & 0.471     & \textbf{0.505}       & 0.395    \\  &
01\_Powdery\_Mildew           & 0.089     & 0.992${}^{\dagger}$   & \textbf{0.706}     & 0.495       & 0.438    \\  &
02\_Gray\_Mold                & 0.115     & 0.972${}^{\dagger}$   & \textbf{0.757}     & 0.558       & 0.689    \\  &
03\_Anthracnose               & 0.747     & 0.956${}^{\dagger}$   & \textbf{0.973}     & 0.862       & 0.876    \\  &
08\_Downy\_Mildew             & 0.456     & 0.999${}^{\dagger}$   & \textbf{0.662}     & 0.552       & 0.467    \\  &
09\_Corynespora\_Leaf\_Spot   & 0.260     & 0.981${}^{\dagger}$   & \textbf{0.736}     & 0.519       & 0.562    \\  &
17\_Gummy\_Stem\_Blight       & 0.502     & 0.863${}^{\dagger}$   & 0.558     & 0.388       & \textbf{0.601}    \\  &
20\_Bacterial\_Spot           & 0.452     & 0.836${}^{\dagger}$   & \textbf{0.595}     & 0.359       & 0.470    \\  &
22\_CCYV                      & 0.238     & 0.921${}^{\dagger}$   & \textbf{0.924}     & 0.795       & 0.645    \\  &
23\_Mosaic\_diseases          & 0.178     & 0.889${}^{\dagger}$   & \textbf{0.718}     & 0.698       & 0.631    \\  &
24\_MYSV                      & 0.211     & 0.980${}^{\dagger}$   & \textbf{0.734}     & 0.511       & 0.492    \\
\cmidrule(lr){2-7}  &
\multicolumn{1}{c}{$R$}       & 0.232     & 0.944${}^{\dagger}$   & \textbf{0.717}     & 0.564       & 0.553   \\

\midrule

% tomato
\multirow{13}{*}{tomato} &
00\_HEALTHY                   & 0.422     & \textbf{0.749}   & 0.996${}^{\dagger}$     & 0.743       & 0.609    \\  &
01\_Powdery\_Mildew           & 0.542     & \textbf{0.699}   & 0.998${}^{\dagger}$     & 0.658       & 0.457    \\  &
02\_Gray\_Mold                & 0.448     & 0.707   & 0.999${}^{\dagger}$     & \textbf{0.844}       & 0.489    \\  &
05\_Cercospora\_Leaf\_Mold    & 0.585     & 0.902   & 0.992${}^{\dagger}$     & \textbf{0.784}       & 0.705    \\  &
06\_Leaf\_Mold                & 0.374     & 0.828   & 0.943${}^{\dagger}$     & \textbf{0.833}       & 0.736    \\  &
07\_Late\_Blight              & 0.452     & 0.436   & 0.985${}^{\dagger}$     & \textbf{0.654}       & 0.462    \\  &
10\_Corynespora\_Target\_Spot & 0.394     & 0.427   & 0.823${}^{\dagger}$     & \textbf{0.556}       & 0.471    \\  &
19\_Bacterial\_Wilt           & 0.158     & \textbf{0.922}   & 0.999${}^{\dagger}$     & 0.780       & 0.503    \\  &
21\_Bacterial\_Canker         & 0.744     & \textbf{0.954}   & 0.987${}^{\dagger}$     & 0.949       & 0.815    \\  &
27\_ToMV                      & 0.610     & \textbf{0.793}   & 0.823${}^{\dagger}$     & 0.780       & 0.637    \\  &
28\_ToCV                      & 0.514     & \textbf{0.736}   & 0.968${}^{\dagger}$     & 0.584       & 0.487    \\  &
29\_Yellow\_Leaf\_Curl        & 0.451     & 0.665   & 0.990${}^{\dagger}$     & \textbf{0.715}       & 0.613    \\
\cmidrule(lr){2-7}  &
\multicolumn{1}{c}{$R$}       & 0.468     & 0.743   & 0.966${}^{\dagger}$     & \textbf{0.745}       & 0.574    \\

\midrule

% eggplant
\multirow{8}{*}{eggplant} &
00\_HEALTHY                   & 0.483     & \textbf{0.936}   & 0.931     & 0.997${}^{\dagger}$       & 0.675    \\  &
01\_Powdery\_Mildew           & 0.312     & \textbf{0.979}   & 0.937     & 0.948${}^{\dagger}$       & 0.663    \\  &
02\_Gray\_Mold                & 0.819     & 0.686   & \textbf{0.883}     & 0.917${}^{\dagger}$       & 0.809    \\  &
06\_Leaf\_Mold                & 0.904     & \textbf{0.873}   & 0.756     & 0.757${}^{\dagger}$       & 0.556    \\  &
11\_Leaf\_Spot                & 0.447     & 0.940   & \textbf{0.948}     & 0.994${}^{\dagger}$       & 0.958    \\  &
18\_Verticillium\_Wilt        & 0.640     & \textbf{0.897}   & 0.868     & 0.914${}^{\dagger}$       & 0.693    \\  &
19\_Bacterial\_Wilt           & 0.469     & \textbf{0.951}   & 0.767     & 0.962${}^{\dagger}$       & 0.489    \\
\cmidrule(lr){2-7}  &
\multicolumn{1}{c}{$R$}       & 0.544     & \textbf{0.909}   & 0.876     & 0.932${}^{\dagger}$       & 0.704    \\

\midrule

% strawberry
\multirow{5}{*}{strawberry} &
00\_HEALTHY                   & 0.585     & \textbf{0.792}   & 0.668     & 0.707       & 0.880${}^{\dagger}$    \\  &
01\_Powdery\_Mildew           & 0.968     & \textbf{0.977}   & 0.973     & 0.933       & 0.978${}^{\dagger}$    \\  &
03\_Anthracnose               & 0.652     & 0.838   & \textbf{0.945}     & 0.932       & 0.956${}^{\dagger}$    \\  &
15\_Fusarium\_Wilt            & 0.641     & \textbf{0.975}   & 0.913     & 0.940       & 0.948${}^{\dagger}$    \\
\cmidrule(lr){2-7}   &
\multicolumn{1}{c}{$R$}       & 0.714     & \textbf{0.896}   & 0.864     & 0.868       & 0.925${}^{\dagger}$    \\

\bottomrule \\
\multicolumn{6}{l}{${}^{\dagger}$ The results are for reference only, as the training data for $M_E$ includes some of the training data for $M_C$.} \\
\multicolumn{6}{l}{Bolded values indicate the best values excluding reference results.}
\end{tabular}
\end{table*}

%% file: 07_limitation.tex
To evaluate whether the similarity ($S$) obtained in this study is appropriate as DDD, we measure correlation ($R$) at $\alpha=1.0$ between $\Tilde{P}$, calculated based on $P$ obtained using $M_C$, and $S$, calculated from $M_E$. 
Therefore, it can only evaluate data within the same class (disease) in a dataset.
In addition, its validity is affected by the performance of the classifier $M_C$.
In this experiment, we only compared the use of a single crop type for training $M_E$, but there is  significant potential for improvement by incorporating tuning with more diverse data from multiple crop types.

%% file: 08_conclusion.tex
This study introduces a novel metric, to quantitatively assess the domain gap between training and test datasets. DDD represents the distance between datasets and reveals the lack of diversity in the training data. The aim is to promote the rapid implementation of strategies to improve the robustness of models, such as incorporating more diverse data based on the results of DDD. As a result of experiments in the plant disease diagnosis task, the distance using low-dimensional representations derived from models trained on additional plant disease datasets that differ from the target crop and disease is more appropriate as a DDD than when only large datasets are used.

%% file: 09_Acknowledgments.tex
This work was supported by the Ministry of Agriculture, Forestry and Fisheries (MAFF) Japan Commissioned project study on "Development of pest diagnosis technology using AI" (JP17935051) and by the Cabinet Office, Public / Private R\&D Investment Strategic Expansion Program (PRISM).

%% file: main.bbl
\begin{thebibliography}{25}
\providecommand{\natexlab}[1]{#1}

\bibitem[{Alvarez-Melis and Fusi(2020)}]{dataset_distance_OTDD}
Alvarez-Melis, D.; and Fusi, N. 2020.
\newblock {Geometric Dataset Distances via Optimal Transport}.

\bibitem[{Atila et~al.(2021)Atila, U{\c{c}}ar, Akyol, and U{\c{c}}ar}]{plant_atila2021plant}
Atila, {\"U}.; U{\c{c}}ar, M.; Akyol, K.; and U{\c{c}}ar, E. 2021.
\newblock {Plant leaf disease classification using EfficientNet deep learning model}.
\newblock \emph{Ecological Informatics}, 61: 101182.

\bibitem[{{Calderon-Ramirez, Saul and Oala, Luis and Torrents-Barrena, Jordina and Yang, Shengxiang and Elizondo, David and Moemeni, Armaghan and Colreavy-Donnelly, Simon and Samek, Wojciech and Molina-Cabello, Miguel A. and López-Rubio, Ezequiel}(2023)}]{dataset_distance_DeDiM}
{Calderon-Ramirez, Saul and Oala, Luis and Torrents-Barrena, Jordina and Yang, Shengxiang and Elizondo, David and Moemeni, Armaghan and Colreavy-Donnelly, Simon and Samek, Wojciech and Molina-Cabello, Miguel A. and López-Rubio, Ezequiel}. 2023.
\newblock {Dataset Similarity to Assess Semisupervised Learning Under Distribution Mismatch Between the Labeled and Unlabeled Datasets}.
\newblock \emph{IEEE Transactions on Artificial Intelligence}, 4(2): 282--291.

\bibitem[{Cap et~al.(2022)Cap, Uga, Kagiwada, and Iyatomi}]{leafgan_plant_plant_cap2020leafgan}
Cap, Q.~H.; Uga, H.; Kagiwada, S.; and Iyatomi, H. 2022.
\newblock {LeafGAN: An Effective Data Augmentation Method for Practical Plant Disease Diagnosis}.
\newblock \emph{IEEE Transactions on Automation Science and Engineering}, 19(2): 1258--1267.

\bibitem[{Chechik et~al.(2009)Chechik, Sharma, Shalit, and Bengio}]{contrastive_learning_10.1007/978-3-642-02172-5_2}
Chechik, G.; Sharma, V.; Shalit, U.; and Bengio, S. 2009.
\newblock Large Scale Online Learning of Image Similarity through Ranking.
\newblock In Araujo, H.; Mendon{\c{c}}a, A.~M.; Pinho, A.~J.; and Torres, M.~I., eds., \emph{Pattern Recognition and Image Analysis}, 11--14. Berlin, Heidelberg: Springer Berlin Heidelberg.
\newblock ISBN 978-3-642-02172-5.

\bibitem[{Chen et~al.(2020)Chen, Kornblith, Norouzi, and Hinton}]{contrastive_learning_SimCLR}
Chen, T.; Kornblith, S.; Norouzi, M.; and Hinton, G. 2020.
\newblock {A simple framework for contrastive learning of visual representations}.
\newblock In \emph{International conference on machine learning}, 1597--1607. PMLR.

\bibitem[{Chen et~al.(2019)Chen, Liu, Kira, Wang, and Huang}]{metric_learning_baseline++}
Chen, W.-Y.; Liu, Y.-C.; Kira, Z.; Wang, Y.-C.; and Huang, J.-B. 2019.
\newblock {A Closer Look at Few-shot Classification}.
\newblock In \emph{International Conference on Learning Representations}.

\bibitem[{Deng et~al.(2009)Deng, Dong, Socher, Li, Li, and Fei-Fei}]{ridnik2021imagenet21k}
Deng, J.; Dong, W.; Socher, R.; Li, L.-J.; Li, K.; and Fei-Fei, L. 2009.
\newblock ImageNet: A large-scale hierarchical image database.
\newblock In \emph{2009 IEEE Conference on Computer Vision and Pattern Recognition}, 248--255.

\bibitem[{Elfatimi, Eryigit, and Elfatimi(2022)}]{plant_elfatimi2022beans}
Elfatimi, E.; Eryigit, R.; and Elfatimi, L. 2022.
\newblock {Beans leaf diseases classification using mobilenet models}.
\newblock \emph{IEEE Access}, 10: 9471--9482.

\bibitem[{Fujita et~al.(2018)Fujita, Uga, Kagiwada, and Iyatomi}]{plant_Fujita2018APP}
Fujita, E.~E.; Uga, H.; Kagiwada, S.; and Iyatomi, H. 2018.
\newblock {A Practical Plant Diagnosis System for Field Leaf Images and Feature Visualization}.
\newblock \emph{International Journal of Engineering \& Technology}.

\bibitem[{Hughes, Salath{\'e} et~al.(2015)}]{plant_hughes2015open}
Hughes, D.; Salath{\'e}, M.; et~al. 2015.
\newblock {An open access repository of images on plant health to enable the development of mobile disease diagnostics}.
\newblock \emph{arXiv preprint arXiv:1511.08060}.

\bibitem[{Iwano et~al.(2024)Iwano, Shibuya, Kagiwada, and Iyatomi}]{HODRF_plant_iwano2024hierarchicalobjectdetectionrecognition}
Iwano, K.; Shibuya, S.; Kagiwada, S.; and Iyatomi, H. 2024.
\newblock {Hierarchical Object Detection and Recognition Framework for Practical Plant Disease Diagnosis}.
\newblock arXiv:2407.17906.

\bibitem[{Kanno et~al.(2021)Kanno, Nagasawa, Cap, Shibuya, Uga, Kagiwada, and Iyatomi}]{ppig_plant_9398772}
Kanno, S.; Nagasawa, S.; Cap, Q.~H.; Shibuya, S.; Uga, H.; Kagiwada, S.; and Iyatomi, H. 2021.
\newblock {PPIG: Productive and Pathogenic Image Generation for Plant Disease Diagnosis}.
\newblock In \emph{{2020 IEEE-EMBS Conference on Biomedical Engineering and Sciences (IECBES)}}, 554--559.

\bibitem[{Kawasaki et~al.(2015)Kawasaki, Uga, Kagiwada, and Iyatomi}]{plant_kawasaki2015basic}
Kawasaki, Y.; Uga, H.; Kagiwada, S.; and Iyatomi, H. 2015.
\newblock {Basic study of automated diagnosis of viral plant diseases using convolutional neural networks}.
\newblock In \emph{Advances in Visual Computing: 11th International Symposium, ISVC 2015, Las Vegas, NV, USA, December 14-16, 2015, Proceedings, Part II 11}, 638--645. Springer.

\bibitem[{Mohanty, Hughes, and Salath{\'e}(2016)}]{plant_mohanty}
Mohanty, S.~P.; Hughes, D.~P.; and Salath{\'e}, M. 2016.
\newblock {Using deep learning for image-based plant disease detection}.
\newblock \emph{Frontiers in plant science}, 7: 1419.

\bibitem[{Narayanan et~al.(2022)Narayanan, Krishnan, Robinson, Julie, Vimal, Saravanan, and Kaliappan}]{plant_narayanan2022banana}
Narayanan, K.~L.; Krishnan, R.~S.; Robinson, Y.~H.; Julie, E.~G.; Vimal, S.; Saravanan, V.; and Kaliappan, M. 2022.
\newblock {Banana plant disease classification using hybrid convolutional neural network}.
\newblock \emph{Computational Intelligence and Neuroscience}, 2022(1): 9153699.

\bibitem[{Radford et~al.(2021)Radford, Kim, Hallacy, Ramesh, Goh, Agarwal, Sastry, Askell, Mishkin, Clark et~al.}]{contrastive_learning_CLIP}
Radford, A.; Kim, J.~W.; Hallacy, C.; Ramesh, A.; Goh, G.; Agarwal, S.; Sastry, G.; Askell, A.; Mishkin, P.; Clark, J.; et~al. 2021.
\newblock {Learning transferable visual models from natural language supervision}.
\newblock In \emph{International conference on machine learning}, 8748--8763. PMLR.

\bibitem[{Ramcharan et~al.(2017)Ramcharan, Baranowski, McCloskey, Ahmed, Legg, and Hughes}]{plant_ramcharan2017deep}
Ramcharan, A.; Baranowski, K.; McCloskey, P.; Ahmed, B.; Legg, J.; and Hughes, D.~P. 2017.
\newblock {Deep learning for image-based cassava disease detection}.
\newblock \emph{Frontiers in plant science}, 8: 1852.

\bibitem[{Shibuya et~al.(2021)Shibuya, Cap, Nagasawa, Kagiwada, Uga, and Iyatomi}]{21k_plant_shibuya2021validation}
Shibuya, S.; Cap, Q.~H.; Nagasawa, S.; Kagiwada, S.; Uga, H.; and Iyatomi, H. 2021.
\newblock {Validation of prerequisites for correct performance evaluation of image-based plant disease diagnosis using reliable 221k images collected from actual fields}.
\newblock In \emph{AI for Agriculture and Food Systems}.

\bibitem[{Sohn(2016)}]{metric_learning_n-class}
Sohn, K. 2016.
\newblock {Improved deep metric learning with multi-class N-pair loss objective}.
\newblock In \emph{Proceedings of the 30th International Conference on Neural Information Processing Systems}, NIPS'16, 1857–1865. Red Hook, NY, USA: Curran Associates Inc.
\newblock ISBN 9781510838819.

\bibitem[{Tan and Le(2021)}]{tan2021efficientnetv2}
Tan, M.; and Le, Q. 2021.
\newblock {Efficientnetv2: Smaller models and faster training}.
\newblock In \emph{International conference on machine learning}, 10096--10106. PMLR.

\bibitem[{Toda and Okura(2019)}]{plant_toda2019convolutional}
Toda, Y.; and Okura, F. 2019.
\newblock {How convolutional neural networks diagnose plant disease}.
\newblock \emph{Plant Phenomics}.

\bibitem[{Wang, Sun, and Wang(2017)}]{plant_wang2017automatic}
Wang, G.; Sun, Y.; and Wang, J. 2017.
\newblock {Automatic image-based plant disease severity estimation using deep learning}.
\newblock \emph{Computational intelligence and neuroscience}, 2017(1): 2917536.

\bibitem[{Wang et~al.(2014)Wang, Song, Leung, Rosenberg, Wang, Philbin, Chen, and Wu}]{metric_learning_triplet}
Wang, J.; Song, Y.; Leung, T.; Rosenberg, C.; Wang, J.; Philbin, J.; Chen, B.; and Wu, Y. 2014.
\newblock {Learning Fine-Grained Image Similarity with Deep Ranking}.
\newblock In \emph{Proceedings of the 2014 IEEE Conference on Computer Vision and Pattern Recognition}, CVPR '14, 1386–1393. USA: IEEE Computer Society.
\newblock ISBN 9781479951185.

\bibitem[{Xie et~al.(2020)Xie, Luong, Hovy, and Le}]{Self-training}
Xie, Q.; Luong, M.-T.; Hovy, E.; and Le, Q.~V. 2020.
\newblock Self-Training With Noisy Student Improves ImageNet Classification.
\newblock In \emph{Proceedings of the IEEE/CVF Conference on Computer Vision and Pattern Recognition (CVPR)}.

\end{thebibliography}
